\documentclass[10pt,twocolumn,letterpaper]{article}

\usepackage{iccv}
\usepackage{times}
\usepackage{epsfig}

\usepackage{graphicx, amsfonts, amsmath, amssymb, booktabs, caption, subcaption, multirow, overpic, textpos, cuted, bm, float}
\usepackage{lipsum}         
\usepackage{enumitem}
\usepackage[table]{xcolor}
\usepackage[T1]{fontenc}
\usepackage[british, american]{babel}
\usepackage[pagebackref=true, breaklinks=true, letterpaper=true, colorlinks,
            citecolor=citecolor, linkcolor=linkcolor, bookmarks=false]{hyperref}
\definecolor{citecolor}{HTML}{0071BC}
\definecolor{linkcolor}{HTML}{ED1C24} 

\usepackage[capitalize]{cleveref}
\crefname{section}{Sec.}{Secs.}
\Crefname{section}{Section}{Sections}
\Crefname{table}{Table}{Tables}
\crefname{table}{Tab.}{Tabs.}

\iccvfinalcopy 

\ificcvfinal\fi

\newlength\savewidth

\renewcommand{\paragraph}[1]{\vspace{1.25mm}\noindent\textbf{#1}}

\newcolumntype{x}[1]{>{\centering\arraybackslash}p{#1pt}}
\newcolumntype{y}[1]{>{\raggedright\arraybackslash}p{#1pt}}
\newcolumntype{z}[1]{>{\raggedleft\arraybackslash}p{#1pt}}

\newcommand{\app}{\raise.17ex\hbox{$\scriptstyle\sim$}}

\definecolor{deemph}{gray}{0.6}

\definecolor{baselinecolor}{gray}{.9}

\newcommand{\mysmall}[1]{\scriptsize{\color{gray}{#1}}}

\newcommand*{\affaddr}[1]{#1} 
\newcommand*{\affmark}[1][*]{\textsuperscript{#1}}

\begin{document}

\title{MODA: Mapping-Once Audio-driven Portrait Animation with Dual Attentions}

\author{Yunfei Liu\affmark[1] \qquad Lijian Lin\affmark[1]  \qquad Fei Yu\affmark[2] \qquad Changyin Zhou\affmark[2] \qquad  Yu Li\affmark[1*] \\
\affaddr{\affmark[1]International Digital Economy Academy (IDEA) \qquad \affmark[2]Vistring Inc.}\\
\small \url{https://tinyurl.com/iccv23-moda}
}

\ificcvfinal\thispagestyle{empty}\fi

\twocolumn[{%
\renewcommand\twocolumn[1][]{#1}%
\maketitle
\begin{center}
    \centering
    \captionsetup{type=figure}
    \includegraphics[width=0.99\textwidth]{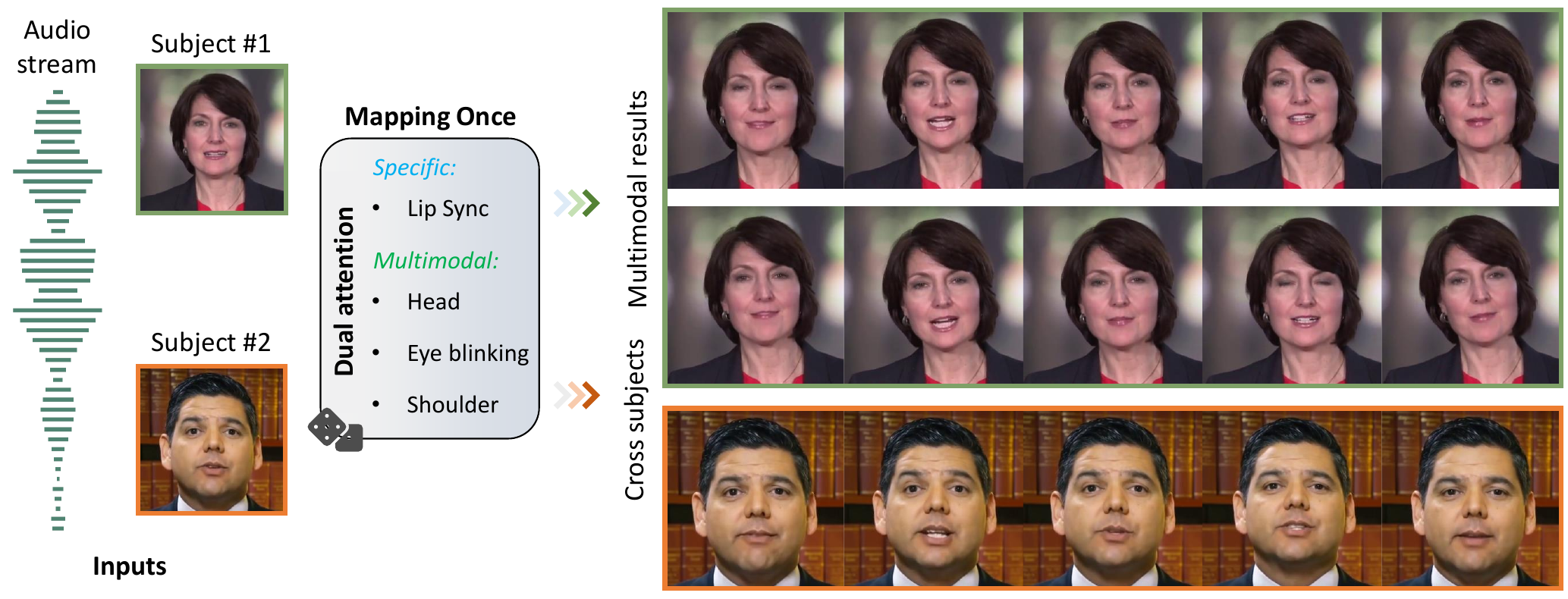}
    \captionof{figure}{We propose a mapping-once system with dual-attention for multimodal and high-fidelity portrait video animation.}
    \label{fig:teaser}
\end{center}%
}]

\begin{abstract}
   Audio-driven portrait animation aims to synthesize portrait videos that are conditioned by given audio. Animating high-fidelity and multimodal video portraits has a variety of applications. Previous methods have attempted to capture different motion modes and generate high-fidelity portrait videos by training different models or sampling signals from given videos. However, lacking correlation learning between lip-sync and other movements (\eg, head pose/eye blinking) usually leads to unnatural results. In this paper, we propose a unified system for multi-person, diverse, and high-fidelity talking portrait generation. Our method contains three stages, \ie, 1) Mapping-Once network with Dual Attentions (MODA) generates talking representation from given audio. In MODA, we design a dual-attention module to encode accurate mouth movements and diverse modalities. 2) Facial composer network generates dense and detailed face landmarks, and 3) temporal-guided renderer syntheses stable videos. Extensive evaluations demonstrate that the proposed system produces more natural and realistic video portraits compared to previous methods.

\end{abstract}

\section{Introduction}
\label{sec:intro}

Given an input audio, talking portrait animation is to synthesize video frames of a person whose poses and expressions are synchronized with the audio signal~\cite{chen2020talking,lele2019Hierarchical,chung2017lip,kr2019towards}. This audio-driven portrait video generation task has gained increasing attention recently and has a wide range of applications in digital avatars, gaming, telepresence, virtual reality (VR), video production \etc. Conventional portrait video generation consumes intensive labor and time during setting up the background, make-up, lighting, shooting, and editing. Moreover, a re-shot is always required when there exists new textual content. In contrast, audio-driven talking video generation is more convenient and attractive which only requires a new audio clip to render a new video.    

Previous methods~\cite{gafni20214d_avatar,prajwal2020lip,zhou2021pose} try to learn the correspondence between audio and frames. However, these methods usually ignore the head pose as it is hard to separate head posture from facial movement. Many 3D face reconstruction algorithm-based and GAN-based~\cite{Goodfellow2014GAN} methods estimate intermediate representations, such as 3D face shapes~\cite{faceformer2022,zhang2022sadtalker}, 2D landmarks~\cite{lu2021LSP,zhou2020MakeItTalk}, or face expression parameters~\cite{zhang2021facial}, to assist the generation process. However, such sparse representations usually lost facial details, leading to over-smooth~\cite{ye2023geneface}. Recently, the neural radiance field (NeRF)~\cite{guo2021adnerf,ye2023geneface} has been widely applied in talking head generation for high-fidelity results. However, the implicit neural representation is hard to interpret and control. In addition, these methods are usually person-specific and require extra training or adaptation time for different persons. 

Although quite a number of attempts and progresses have been made in recent years, it is still challenging to generate realistic and expressive talking videos. As humans are extremely sensitive to identifying the artifacts in the synthesized portrait videos, it sets a very high standard for this technique to become applicable. We summarize the following key points that affect human perceptions: 1) \textbf{Correctness}. The synthesized talking portrait video should be well synchronized with the driven audio. 2) \textbf{Visual quality}. The synthesized video should have high resolution and contain fine detail components. 3) \textbf{Diversity}. Besides the lip motion needing to be exactly matched to the audio content, the motion of other components like eye blinking and head movement are not deterministic. They should move naturally as a natural human does.

To achieve these goals, previous approaches either map the mouth landmarks and the head pose separately by learning different sub-networks~\cite{lu2021LSP,zhang2022sadtalker}, or only model the mouth movement while the head pose is obtained from the existing video~\cite{prajwal2020lip, zhou2021pose}. However, lacking correlation learning between lip-sync and other movements usually leads to unnatural results. In this paper, we propose a \textbf{m}apping-\textbf{o}nce network with \textbf{d}ual \textbf{a}ttentions (MODA), which is a unified architecture to generate diverse representations for a talking portrait, simplifying the computational steps. In order to combine synchronization and diversity of the talking portrait generation, we carefully design a dual-attention module to learn deterministic mappings  (\ie, the accurate mouth movements driven by audio) and probabilistic sampling (\ie, the diverse head pose/eye blinking from time-to-time), respectively. 
To summarize, our contributions can be listed as follows:

\begin{itemize}[itemsep=2pt,topsep=0pt,parsep=0pt]
    \item We propose a talking portrait system that generates multimodal photorealistic portrait videos with accurate lip motion. Comprehensive evaluations demonstrate our system can achieve state-of-the-art performance.
    \item We propose a unified mapping-once with dual attention (MODA) network for generating portrait representation from subject conditions and arbitrary audio. 
    \item We propose 3 technical points for taking portrait generation: 1) A transformer-based dual attention module for generating both specific and diverse representations. 2) A facial composer network to get accurate and detailed facial landmarks. 3) A temporally guided renderer to synthesize videos with both high quality and temporal stabilization. 
\end{itemize}

\begin{figure*}[htbp]
  \centering
   \includegraphics[width=1\linewidth]{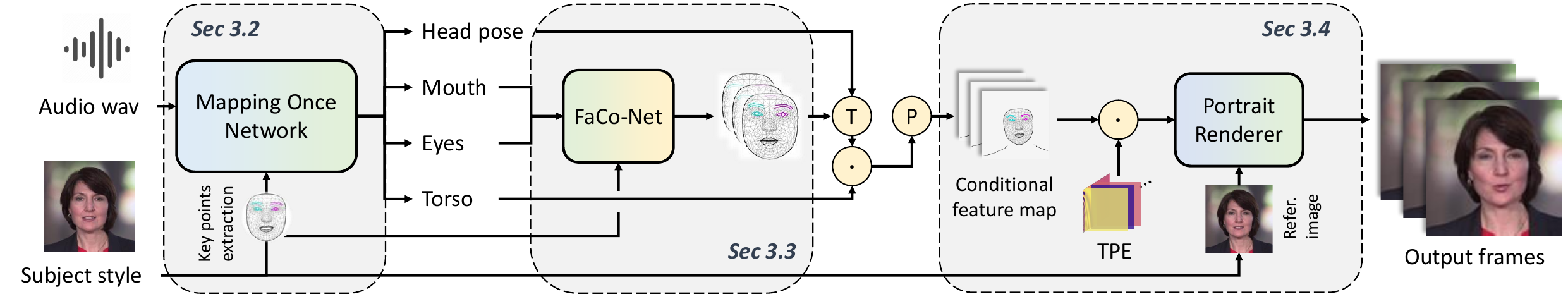}
   \caption{The proposed method is a three-stage system. Given the subject figure and arbitrary audio, the proposed system generates audio-driven video. Here \textcircled{\scriptsize{T}} denotes rigid transformation from canonical space to camera coordinate via head pose, $\odot$ denotes concatenation. \textcircled{\scriptsize{P}} is projection from 3D space to image coordinate. FaCo-Net is a facial composer network, which will be introduced in \cref{sec:refiner}.}
   \label{fig:method_overview}
   \vspace{-0.5cm}
\end{figure*}

\section{Related Works}
\label{sec:related}

\noindent \textbf{Audio-driven portrait animation.}
Talking heads and facial animation are research hot-spots in the computer vision community. Extensive approaches~\cite{edwards2016jali,zhou2018visemenet} explore audio-driven mouth animation and audio-driven facial animation. We focus on animating a portrait in this work. 
Many methods~\cite{chung2017lip,edwards2016jali,prajwal2020lip,schreer2008real,zhu2018high} aim to find the correspondence between audio and frames. A large number of technologies (such as flow-learning~\cite{ji2022eamm,wang2021audio2head,zhang2021HDTF}, memory bank~\cite{park2022synctalkface,tang2022memories}, \etc) are explored for the correctness of talking head generation.
However, these methods usually ignore the head pose, torso motion, and eye blinking, which are essential for a natural talking portrait generation. 
To generate diverse talking heads, recent methods~\cite{Lahiri_2021_CVPR,zhou2021pose} propose to embed other modalities to control emotions or head pose. However, these methods usually require additional inputs.

Recently, neural radiance field (NeRF)~\cite{guo2021adnerf} has been widely applied in 3D-related tasks as it can accurately reproduce complex scenes with implicit neural representation. Several works~\cite{guo2021adnerf,yao2022dfa,liu2022semantic,ye2023geneface} leverage NeRF to represent faces with audio features as conditions. Despite the high-quality results achieved, the motion of generated results is usually unnatural. Besides, the learning and inference processes are time-consuming.  More recently, some diffusion-based methods~\cite{shen2023difftalk,stypulkowski2023diffused} are proposed to generate talking heads. However, their speed is limited by a large number of sampling steps in the diffusion process. 
Some methods are based on the 3D face reconstruction and GAN~\cite{karras2017audio,lu2021LSP,yi2020audio}. They estimate intermediate representations such as 2D landmarks~\cite{wang2020mead,zakharov2019few,zhou2020MakeItTalk}, 3D face shapes~\cite{karras2017audio,thies2020neural} or facial expression parameters~\cite{wang2021anyonenet,zhang2022sadtalker}, to assist the generation process. Unfortunately, such sparse representation usually lost facial details. In this paper, we propose to learn dense facial landmarks and upper body points through a unified framework for talking portrait generation. The intermediate representation contains facial details and other movements, which can be interpreted and controlled easily. 

\noindent \textbf{Transformers in audio-driven tasks.}
Transformer~\cite{vaswani2017attention} is a strong alternative to both RNN and CNN. Researchers find it works well in multimodal scenarios. We refer readers to the comprehensive survey~\cite{khan2021transformers} for further information. Some recent works adopt transformers to generate results from different modalities, such as audio-to-text, language translation, music-to-dance, \etc. The most related work is FaceFormer~\cite{faceformer2022}, which is a speech-driven 3D facial animation approach. They proposed two types of bias for the transformer to better align audio and 3D face animation.

\noindent \textbf{Vision-based facial reenactment.}
Video-based facial reenactment is another technique related to audio-driven animation~\cite{siarohin2019first,schreer2008real}. There are many works to reenact faces with different techniques, such as adversarial learning, few-shot learning, or even one-shot facial animation. They usually adopt pre-defined facial landmarks or in an unsupervised scheme. In another aspect, image-to-image translation (I2I) methods~\cite{Liu2017UnsupervisedImageTranslation,Yi2017Dualgan} have also demonstrated impressive performance in converting images from one domain to another. 
However, single-frame renderers~\cite{Liu2017UnsupervisedImageTranslation} ignore the temporal relations among video frames, leading to color jitters or unnatural background shakes in the final results.
\cite{wang2018vid2vid} proposes to use RNN~\cite{medsker2001recurrent} to capture the temporal relations among the input conditions, which generates stabilized results. However, these methods have difficulties in training~\cite{pascanu2013difficulty}.
In this paper, we find an alternative way to embed temporal information into I2I. Simply using temporal positional embedding~\cite{han2022survey} as an input condition, our method can achieve natural and stabilized results.


\begin{figure*}[htbp]
  \centering
   \includegraphics[width=1\linewidth]{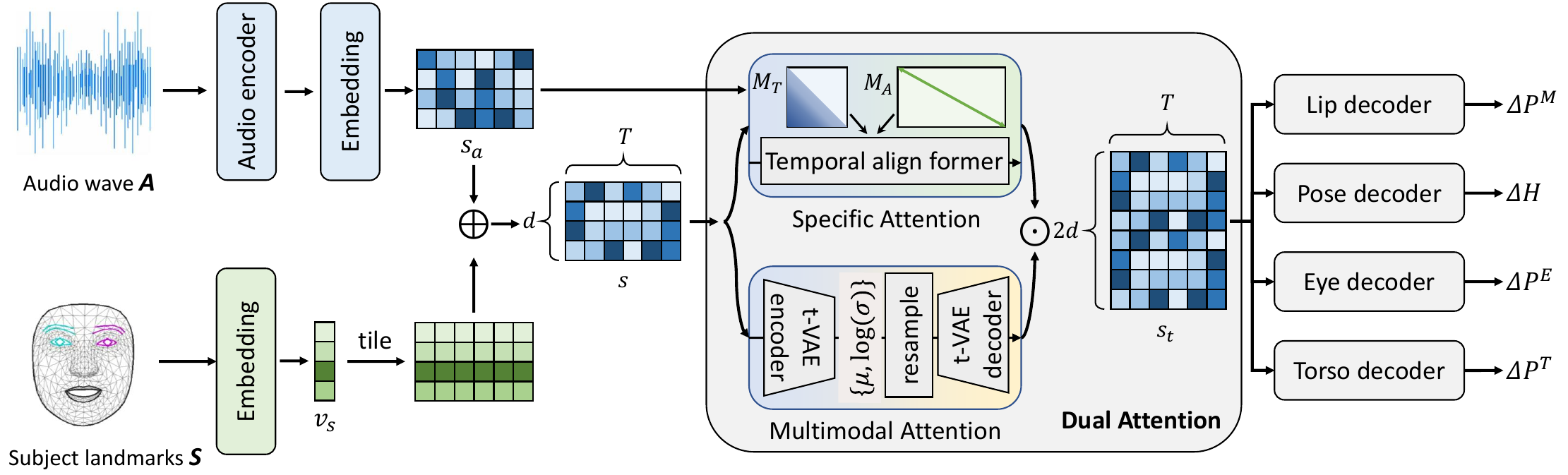}
   \caption{Architecture of MODA network. Given an audio and subject condition, MODA generates four types of motions within a single forward process. $\oplus$ denotes element-wise addition and $\odot$ is concatenation.}
   \label{fig:MOTP}
   \vspace{-0.4cm}
\end{figure*}

\section{Methodology}
\label{sec:method}
We present a talking portrait system for high-fidelity portrait video generation with accurate lip motion and multi-modal motions, including head pose, eye blinking, and torso movements. 
The overall pipeline of this system is illustrated in \cref{fig:method_overview}. It contains three stages, 1) given the driven audio and conditioned subjects, mapping once talking portrait network with dual attentions (MODA) generates multimodal and correct semantic portrait components, 2) in the next, the facial composer network combines the facial components together and adds details for dense facial vertices, and 3) finally, a portrait renderer with temporally positional embedding (TPE) syntheses high-fidelity and stable videos.

\subsection{Task Definition}
\label{sec:task_def}
In this section, we give the definition of the talking portrait task, which is to formulate a sequence-to-sequence translation manner~\cite{beltagy2020longformer} from talking portrait videos. Specifically, given a $T$-length audio sequence $\mathbf{A} = \{a_0, a_1, \dots, a_T\}$ with audio sampling rate $r$, a talking portrait method aims to map it into the corresponding video clip $\mathbf{V} = \{I_0, I_1, \dots, I_K\}$ with $f$ frame-per-second (FPS), where $K = \lfloor fT/r\rfloor$. Since the data dimension of $\mathbf{V}$ is much larger than $\mathbf{A}$, many researchers propose to generate $\mathbf{V}$ progressively and introduce many types of intermediate representation $\mathbf{R}$. 
To make the generated $\mathbf{V}$ look natural, the constraint on $\mathbf{R}$ is critical. In previous audio-driven face animation approaches, $\mathbf{R}$ typically represents one type of face information, such as facial landmarks~\cite{lu2021LSP,zhou2020MakeItTalk} or head pose~\cite{zhang2022sadtalker}. To better represent a talking portrait, we define $\mathbf{R}$ as the union of different portrait descriptors, \ie, $\mathbf{R} = \{P^M, P^E, P^F, H, P^T\}$, where the elements of $\mathbf{R}$ are defined as follows,

\begin{enumerate}[itemsep=2pt,topsep=0pt,parsep=0pt]
    \item Mouth points $P^M \in \mathbb{R}^{40 \times 3}$. They have 40 points for representing mouth animation.
    \item Eyes points $P^E \in \mathbb{R}^{60 \times 3}$. They consist of eye and eyebrow points, which control eye blinking. 
    \item Facial points $P^F \in \mathbb{R}^{478 \times 3}$. They contain dense facial 3D points for recording expression details.
    \item Head pose $H \in \mathbb{R}^{6}$. It contains head rotations ($\theta, \phi, \psi$ in Euler angle) and head transposes ($x, y, z$ in Euclidean space).
    \item Torso points $P^T \in \mathbb{R}^{18\times 3}$. They contain 18 points and each side of the shoulder is described by 9 points.
\end{enumerate}

Note that $P^M, P^E, \text{and~} P^F$ are in canonical space for the convenience of face alignment. The process of talking portrait can be rewritten as $\mathbf{A} \to \mathbf{R} \to \mathbf{V}$. We design corresponding networks for these stages, respectively. The details are provided in the following subsections.

\subsection{Mapping-Once Network with Dual Attentions}
\label{sec:audio2feature}

\noindent \textbf{Mapping-once architecture.}
Given the driven audio $\mathbf{A}$ and subject condition $\mathbf{S}$, MODA aims to map them into $\mathbf{R}$ (consists of lip movement, eye blinking, head pose, and torso) with a single forward process. 
As illustrated in \cref{fig:MOTP}, the network in the first step contains three parts, \ie, 1) two encoders for encoding audio features and extracting subject style, respectively, 2) a dual-attention module for generating diverse but accurate motion features, and 3) four tails for different motion synthesis.
We first extract contextual features of the audio signal by Wav2Vec~\cite{2019wav2vec}. In the next, the extracted feature is projected into $s_a \in \mathbb{R}^{d\times T}$ via a multilayer perceptron (MLP), where $d$ is the feature dimension for one frame and $T$ denotes the number of frames of the generated video.
To model different speaker styles, we take the facial vertices of the conditioned subject as input. Then those vertices are projected to a $d$-dimensional vector $v_s$ as the subject style code. Here the embedding layer is implemented by MLP.
Next, $s_a$ and $v_s$ are combined as:

\begin{equation}
    s = s_a \oplus \text{tile}(v_s), 
\end{equation}
where $s$ is the combined feature, $\oplus$ is dimension-wise addition. Then the dual-attention module (DualAttn) takes $s$, $s_a$ as input, and yields a temporally contextual version $s_t$, 
\begin{equation}
    s_t = \text{DualAttn}(s, s_a).
\end{equation}
Next, we adopt 4 MLPs to decode the movements of lips $P^M$, head pose $H$, eye blinking $P^E$, and torso $P^S$, respectively. For each downstream task $X$, the computational process can be formulated as follows,
\begin{equation}
    \Delta X =  \Phi^X(s_t),
\end{equation}
where $\Phi(\cdot)$ denotes an MLP and $\Delta X = X - \overline{X}$, $\overline{X}$ is extracted from referred subject image.

\noindent \textbf{Dual-attention module.}
The talking portrait generation task is highly ill-posed since it requires generating multimodal results from limited-driven information. To solve this, we propose a dual-attention module that disentangles this task into a \textit{specific mapping} and a \textit{probabilistic mapping} problem. Specifically, this module generates 1) the temporally aligned feature for specific mapping between audio and lip movements, as well as 2) the temporally correlated feature for probabilistic mapping between audio and other movements of the talking portrait. To this end, we first design two sub-modules to learn these two different features, respectively. Then we fuse these two features via time-wise concatenation.

In detail, we propose a \textit{specific attention} branch (SpecAttn) to capture the temporally aligned attention $s_{sa}$ between $s$ and audio feature $s_a$. 
Inspired by FaceFormer~\cite{faceformer2022}, our SpecAttn is formulated as:
\begin{equation}\label{eq:spec_attn}
\begin{aligned}
    s_{sa} &=  \text{SpecAttn}(s_a, s) \\
           &= \text{softmax}(\frac{\Gamma(s) \cdot s_a^T}{\sqrt{d}} + M_A)\Gamma(s),
\end{aligned}
\end{equation}
where $d$ is the dimension of $s_a$, $\{\cdot\}^T$ indicates the transpose of the input parameter. The alignment bias $M_A(1 \leq i \leq T, 1 \leq j \leq T)$ is represented as:
\begin{equation}
M_A(i, j)=\left\{\begin{array}{ll}
0, & i = j \\
-\infty, & \text { otherwise.}
\end{array}\right.
\end{equation}
Different from FaceFormer which performs cross-attention in an auto-regressive manner, we apply this operation on the entire sequence, which boosts the computation speed $T \times$ faster.
In addition, to capture rich temporal information, we adopt a periodic positional encoding (PPE) and a biased casual self-attention on $s$ (as in~\cite{faceformer2022}):
\begin{equation}
    s' = \Gamma(s) = \text{softmax}(\frac{\text{PPE}(s) \cdot \text{PPE}(s)^T}{\sqrt{d}} + M_T)\text{PPE}(s).
\end{equation}
$M_T$ is a matrix that has negative infinity in the upper triangle to avoid looking at future frames to make current predictions. $M_T$ is defined as:
\begin{equation}
\begin{aligned}
M_T(i,j)=\left\{\begin{array}{ll}
\lfloor (i - j)q \rfloor, &  j \leq i ,\\
-\infty, & \text{otherwise,}
\end{array}\right.
\end{aligned}
\end{equation}
where $q$ is a hyper-parameter for tuning the sequence period. By doing this, the encoded feature $s'$ contains rich spatial-temporal information, which aids the accurate talking portrait generation.

To generate vivid results and avoid the over-smoothing~\cite{ye2023geneface} representations, it is essential to learn the probabilistic mapping between the audio feature and portrait motions. 
We notice that Variational Autoencoder (VAE)~\cite{kingma2014VAE} can model probabilistic synthesis and shows many advanced performances in sequence generation tasks. Therefore, based on an advanced transformer Variational Autoencoder (t-VAE)~\cite{petrovich2021action}, we design a \textit{probabilistic attention} branch to generate diverse results. Formally, given the representation $s$, the probabilistic attention (ProbAttn) aims to generate a diverse feature $s_{pa}$. It first models the distribution of $s$ with learned $\mu$ and $\sigma$ through an encoder (Enc). Then it generates multimodal outputs through a re-sample operation with a decoder (Dec).  
The computational process is 
\begin{equation}
\begin{aligned}
    \mu, \log \sigma &= \Phi^\mu(\text{Enc}(s)), \Phi^\sigma(\text{Enc}(s)), \\
    s_{pa} &= \text{Dec}(x), ~~s.t. ~~x \sim \mathcal{U}(\mu, \sigma),
\end{aligned}
\end{equation}
where $\Phi$ is an MLP. $\mathcal{U}(\mu, \sigma)$ is the Gaussian distribution with mean $\mu$ and variance $\sigma$.
To force ProbAttn to learn diverse motion styles, we add Kullback–Leibler divergence (KLD) loss to constrain the feature from the bottleneck of t-VAE. The KLD loss is defined as follows:
\begin{equation}
    \mathcal{L}_{KLD} = (-\frac{1}{2d_l}\sum^{d_l} (\log \sigma - \mu^2 - \sigma + 1),
\end{equation}
where $d_l$ is the dimension of $\mu$.
Finally, the dual-attention module outputs $s_t = s_{sa}\odot s_{pa}$ for downstream tasks.

\noindent\textbf{Loss functions.} The MODA has four decoders for generating talking portrait-related motions. To learn the mapping from the dual-attention module and four different types of motion, we adopt a multi-task learning scheme for MODA. Specifically, we minimize the $L_1$ distance between the ground-truth displacements and the predicted displacements. The loss can be written as 
\begin{equation}
\begin{aligned}
    \mathcal{L}_{TP} & = \lambda_1|\Delta P^M_{gt} - \Delta P^M| + \lambda_2|\Delta R_{gt} - \Delta R| \\
                     & + \lambda_3|\Delta P^E_{gt} - \Delta P^E| + \lambda_4|\Delta P^S_{gt} - \Delta P^S|,
\end{aligned}
\end{equation}
where $\lambda_1, \lambda_2, \lambda_3, \lambda_4$ are hyper-parameters for balancing the different weights of downstream tasks. $|\cdot|$ is the $L_1$-norm. $\Delta P^*_{gt}$ and $\Delta P^*$ indicate the displacements of the ground truth and the predicted result, respectively.
The total loss function is the sum of $\mathcal{L}_{TP}$ and $\mathcal{L}_{KLD}$, \ie,
\begin{equation}
    \mathcal{L}_{total} = \mathcal{L}_{TP} + \mathcal{L}_{KLD}.
\end{equation}

\begin{table*}[t!]
    \centering
    \caption{Comparisons with state-of-the-art methods. $\dag$ denotes our generated results with size $256\times 256$ through a small renderer. The best results are highlighted in \textbf{bold}. The number with \underline{underline} denotes the second-best result.}
    \label{tab:main_results}

    \small
    \setlength{\tabcolsep}{4.5pt}
    \begin{tabular}{lcccccccccc}
        \toprule
        ~ & \multicolumn{5}{c}{Testset A from LSP~\cite{lu2021LSP}} & \multicolumn{5}{c}{Testset B from HDTF~\cite{zhang2021HDTF}}\\
        
        \cmidrule(lr){2-6} \cmidrule(lr){7-11}
        \text{Method} & $\text{NIQE} \downarrow$ & $\text{LMD-}v \downarrow$ & $\text{LMD} \downarrow$& $\text{Sync} \uparrow$ & $\text{MA} \uparrow$ & $\text{NIQE} \downarrow$ & $\text{LMD-}v \downarrow$ & $\text{LMD} \downarrow$& $\text{Sync} \uparrow$ & $\text{MA} \uparrow$ \\

        \midrule
        
        MakeItTalk~\mysmall{(SIGGRAPH Asia'20~\cite{zhou2020MakeItTalk})} & 7.07 & 2.30 & 2.65 & 3.07 & 0.48 & 8.18 & 1.91 & 2.23 & 3.90  & 0.53 \\ 
        
        Wav2Lip~\mysmall{(MM'20~\cite{prajwal2020lip})} & 7.31 & 1.95 & 1.81 & \textbf{5.58} & 0.64 & 7.83 & 2.08 & 1.97 & \textbf{5.78}  & 0.51 \\ 
        
        Wav2Lip-GAN~\mysmall{(MM'20~\cite{prajwal2020lip})} & 7.24 & 2.11 & 1.83 & 5.47 & 0.62 & 7.77 & 2.01 & 1.98 & 5.78  & 0.51 \\ 
        
        LSP~\mysmall{(SIGGRAPH Asia'21~\cite{lu2021LSP})} & \underline{5.75} & 2.28 & 2.06 & 3.09 & 0.61 & 7.12 & 1.67 & \underline{2.01} & 4.11 & 0.52 \\ 
        
        AD-NeRF~\mysmall{(ICCV'21~\cite{guo2021adnerf})} & 5.81 & 2.89 & 2.77 & 2.98 & 0.41 & - & - & - & - & - \\ 
        
        SadTalker~\mysmall{(CVPR'23~\cite{zhang2022sadtalker})} & 5.80 & 2.51 & 2.31 & 4.14 & 0.56 & 7.07 & 2.43 & 2.37 & 3.96 & 0.51 \\ 
        
        GeneFace~\mysmall{(ICLR'23~\cite{ye2023geneface})} & 6.61 & 2.22 & 2.17 & 3.08 & 0.65 & - & - & - & - & - \\  
        
        \midrule
        Ground Truth (reference) & 5.28 & 0.00 & 0.00 & 4.89 & 1.00  & 6.38 & 0.00 & 0.00 & 6.07  & 1.00 \\ 
        Ours $\dag$ & 5.77 & \textbf{1.74} & \underline{1.51} & 4.52 & \underline{0.70} & \underline{7.05} & \underline{1.60} & 2.04 & 4.34 & \textbf{0.59} \\ 
        Ours & \textbf{5.55} & \underline{1.79} & \textbf{1.50} & 4.48 & \textbf{0.69} & \textbf{6.92} & \textbf{1.59} & \textbf{1.96} & 4.16 & \underline{0.56} \\ 
        
        \bottomrule
    \end{tabular}
    \vspace{-0.5cm}
\end{table*}

\subsection{Facial Composer Network}
\label{sec:refiner}
Given the subject information $\mathbf{S}$, the generated mouth points $P^M$, and eye points $P^E$, the facial composer network (FaCo-Net) aims to composite the refined facial dense landmarks.
The generated facial dense landmarks $P^F = \text{FaCo-Net}(\mathbf{S}, P^M, P^E)$.
FaCo-Net consists of three encoders for consuming those three inputs and a decoder for facial landmarks generation. 
Similar to MODA, the subject encoder projects facial points $\mathbf{S}$ into a style code $\bm{p}_f$. The $P^M$ and $P^E$ are also projected to $\bm{p}_m$ and $\bm{p}_e$, which share the same latent space as $\bm{p}_f$. Next, $P^F = \Psi_c((\bm{p}_m\odot \bm{p}_e)\oplus \bm{p}_f)$, where $\Psi_c$ is a facial dense point decoder.
We adopt a vanilla GAN architecture~\cite{Goodfellow2014GAN} as the backbone of the discriminator ($D$). The FaCo-Net is trained to generate ``realistic'' facial dense points to fool $D$, whereas $D$ is trained to distinguish the generated facial points from ground truths. 
The detailed architectures can be found in the supplementary materials. 
We use LSGAN loss~\cite{Liu2017UnsupervisedImageTranslation} as the adversarial loss to optimize the D:
\begin{equation}
    \mathcal{L}_{Disc}(D) = (z - 1)^2 + \hat{z}^2,
\end{equation}
where $z, \hat{z}$ is the discriminator output when inputting the ground-truth face points $P^F_{gt}$ and the generated $P^F$, respectively.
The loss for the generator is 
\begin{equation}
    \mathcal{L}_G = \mathcal{L}_{GAN}(\text{FaCo-Net}) + \lambda |P^F_{gt} - P^F|,
\end{equation}
where $P^F_{gt}$ is the ground-truth dense face landmarks. $\mathcal{L}_{GAN}(\text{FaCo-Net}) = (\hat{z} -1)^2$ is the adversarial loss, where $\hat{z} = D(P^F)$. The weight $\lambda$ is empirically set to 10. After composition, the facial landmarks $P^F$ are transformed to the camera coordinate via head pose $H$. The transformed facial landmarks and torso points $P^T$ are projected into image space for photorealistic rendering.

\subsection{Portrait Image Synthesis with TPE}
\label{sec:render}
The last stage of our system is a renderer that generates photorealistic facial renderings from previous predictions, as illustrated in \cref{fig:method_overview}. 
Specifically, we design a U-Net-like renderer $G_R$ with TPE to generate both high-fidelity and stable videos. In our experiments, TPE is defined as 
\begin{equation}
\begin{aligned}
    \text{TPE}_{(t, 2i)} & = \sin (t * 2^i / 100), \\
    \text{TPE}_{(t, 2i + 1)} & = \cos (t * 2^i / 100).
\end{aligned}
\end{equation}
$i = 0, 1, ..., 5$ is the dimension and $t$ is the frame index.
Then the rendered result \textit{t}-frame $I_t$ is generated with $G_R$:
\begin{equation}
    I_t = G_R(I^c_t \odot I_r \odot \text{TPE}(t)),
\end{equation}
where $I^c_t$ is the condition image at frame index $t$. $I_r$ is the reference image. The detailed architecture, training, and inference details are provided in our supplementary materials.

\subsection{Implementation Details}
\label{sec:impl}
Our models are trained on PyTorch~\cite{PyTorch} using Adam optimizer with hyper-parameters $(\beta_1, \beta_2) = (0.9, 0.99)$. The learning rate is set to $10^{-4}$ in all experiments. 

We train all of our models on an NVIDIA 3090 GPU. It takes about (30, 2, 6) hours in total, (200, 300, 100) epochs with bath sizes of (32, 32, 4) for our three different stages, respectively. During testing, we select all the models with minimum validation loss. We use a sliding window (window size 300, stride 150) for arbitrary long input audio.


\section{Experiments}
\label{sec:exp}
\subsection{Experimental Setup}
\noindent \textbf{Dataset pre-processing.}
We evaluate our method on two publicly available datasets, \ie, HDTF~\cite{zhang2021HDTF} and Video samples from LSP~\cite{lu2021LSP} (LSP dataset). Each video contains a high-resolution portrait with an audio track. The average video length is 1-5 minutes and we process them at 25 fps. 
We randomly select 80\% of them for training and the remaining videos for evaluation. Specifically, we get 132 videos for training and 32 videos for evaluation. Each video is cropped to keep the face at the center and then resized to $512 \times 512$.
The LSP dataset contains 5 different target sequences of 4 different subjects for training and testing. These sequences span a range of 3-5 minutes. All videos are extracted at 25 fps and the synchronized audio is sampled at $16K$ Hz frequency. We split videos as 80\% / 20\% for training and validation. 

We detect $478$ 3D facial landmarks for all videos using Mediapipe\footnote{\url{https://google.github.io/mediapipe/}}. Then we estimate the head pose $H$ for all videos using method~\cite{guo2020towards}. 
According to these head poses, the 3D facial landmarks are projected to the canonical space through rigid transformation. We extract the 3D mouth points, and eye-related points as $P^M$ and $P^E$ for each frame. 
The torso points are estimated from the boundary of the shoulders, which is detected through the face parsing algorithm\footnote{\url{https://github.com/zllrunning/face-parsing.PyTorch}}. For more data pre-processing details please refer to our supplement materials.

\noindent \textbf{Evaluation metrics.}
We demonstrate the superiority of our method on multiple metrics that are widely involved in the talking portrait field.
To evaluate the correctness of generated mouth, we use mouth landmark distance (LMD) and velocity of mouth landmark distance (LMD-$v$) between generated video and reference video in canonical space. In addition, we also calculate the Insertion-over-Union (IoU) for the overlap between the predicted mouth area and the ground truth area (MA). We use the confidence score from SyncNet (Sync)~\cite{prajwal2020lip} to measure the audio-video synchronization.
Since the result cannot be perfectly aligned with the ground-truth video, we use Natural Image Quality Evaluator (NIQE)~\cite{mittal2012making} as the metric for image quality. NIQE is able to capture the naturalness of image details, it is widely used in blind image quality assessment.

\begin{figure}
  \centering
  \includegraphics[width=\linewidth]{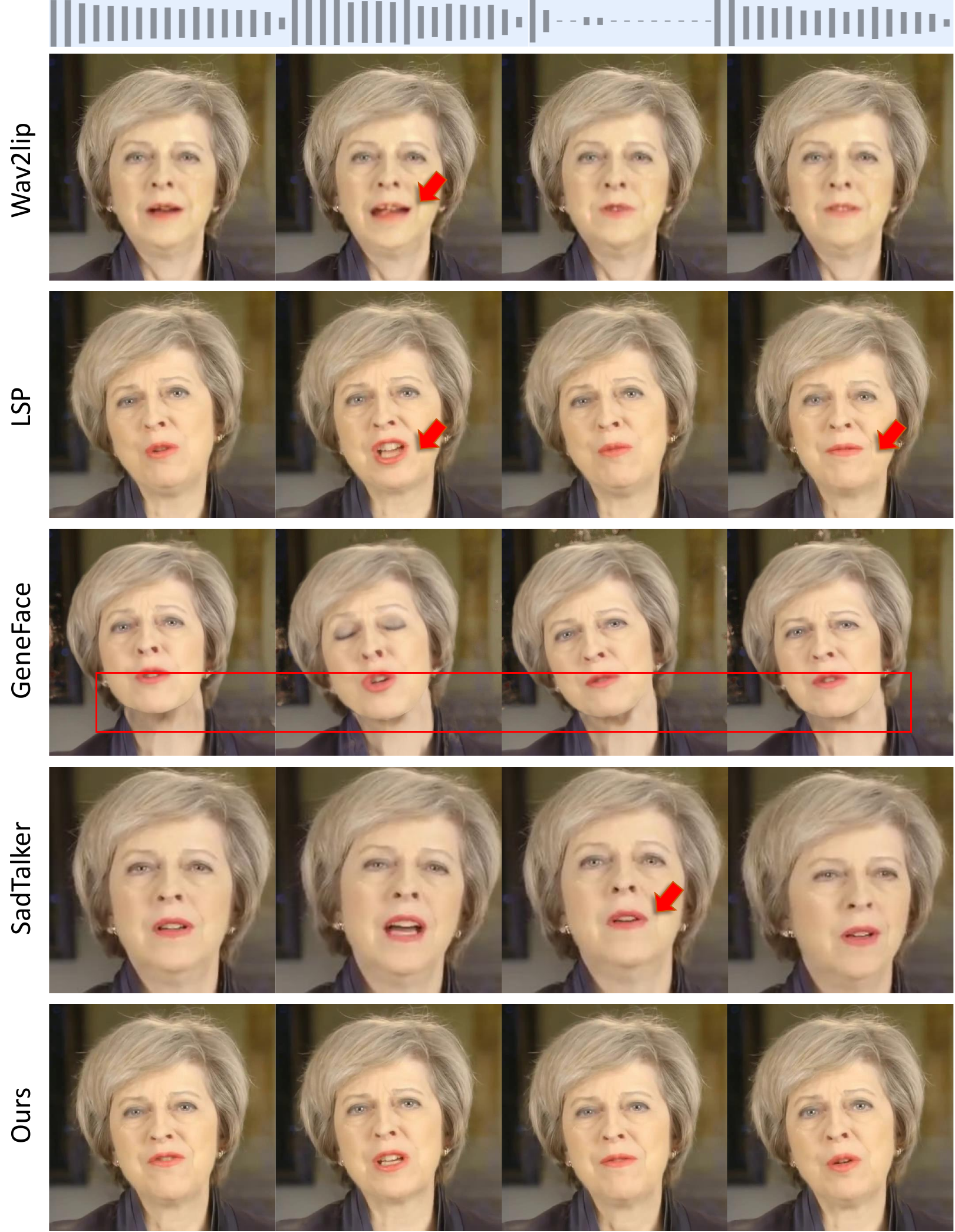}
  \caption{Visual comparison of 5 methods.}
  \label{fig:vis_cmp}
  \vspace{-0.5cm}
\end{figure}

\begin{figure}[h]
  \centering
  \includegraphics[width=\linewidth]{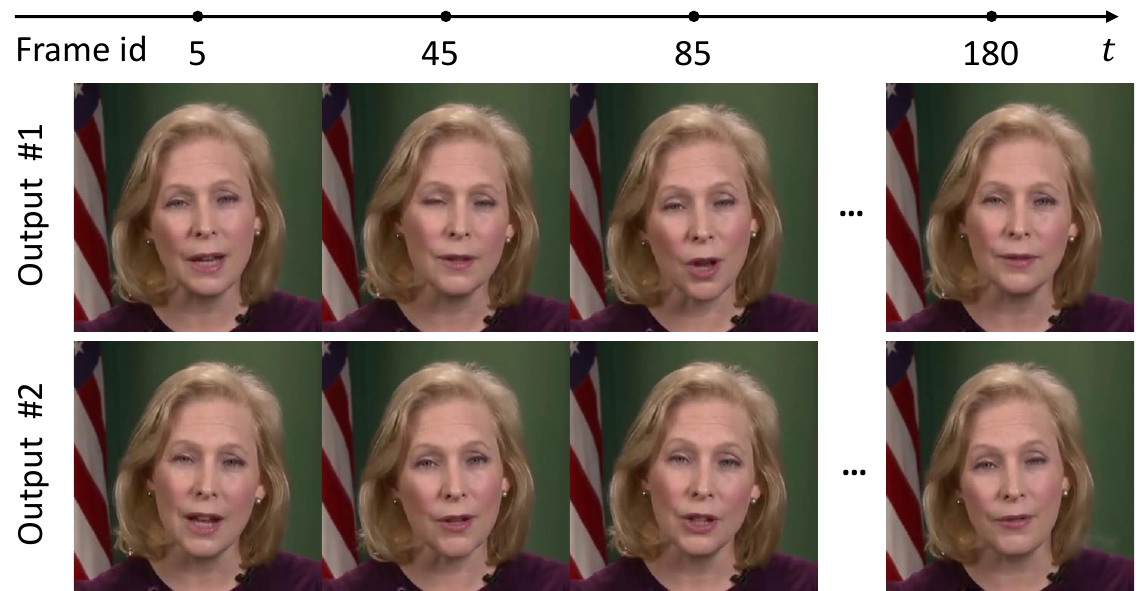}
  \caption{Multimodal results with the same mouth shape.} \vspace{-0.4cm}
  \label{fig:multimodal}
\end{figure}

\subsection{Quantitative Comparison}
We compare our method with several state-of-the-art one-shot talking portrait generation works (LSP~\cite{lu2021LSP}, MakeItTalk~\cite{zhou2020MakeItTalk}, Wav2Lip~\cite{prajwal2020lip}, AD-NeRF~\cite{guo2021adnerf}, SadTalker~\cite{zhang2022sadtalker}, and GeneFace~\cite{ye2023geneface}). For MakeItTalk, Wav2Lip, and SadTalker, the evaluation is performed on their publicly available checkpoint directly. Since these methods only generate low-resolution results, we retrained a small portrait renderer to generate low-resolution results for a fair comparison. The rest methods are retrained on our dataset under the same condition. Note that AD-NeRF and GeneFace are NeRF-based methods that are extremely time-consuming on all videos, we only provide the numerical results on the LSP dataset. 
As shown in \cref{tab:main_results}, the proposed method achieves the best overall video quality (lowest NIQE, $5.25$) and the correctness of audio-lip synchronization (lowest LMD, LMD-$v$, and highest MA). Our method also shows comparable performance with other fully talking-head generation methods in terms of lip-sync score. Please note that a higher sync score is not always lead to better results since it is too sensitive to the audio where unnatural lip movements may get a better score~\cite{zhang2022sadtalker}. 

\begin{table*}[!t]
    \centering
    \caption{User study analyses measured by best-voting percentage. Higher is better.}
    \label{tab:user_study}
    \small
    \setlength{\tabcolsep}{4pt}
    \begin{tabular}{lcccc|cccc}
        \toprule
        ~ & \multicolumn{4}{c|}{Low resolution ($256\times 256$)} & \multicolumn{4}{c}{High resolution ($512\times 512$)}\\
        
        \cmidrule(lr){2-5} \cmidrule(lr){6-9}
        Approach & MakeItTalk\mysmall{~\cite{zhou2020MakeItTalk}} & Wav2lip\mysmall{~\cite{prajwal2020lip}} & SadTalker\mysmall{~\cite{zhang2022sadtalker}} & Ours & LSP\mysmall{~\cite{lu2021LSP}} & AD-NeRF\mysmall{~\cite{guo2021adnerf}} & GeneFace\mysmall{~\cite{ye2023geneface}} & Ours  \\ 
        \midrule
        Lip-sync accuracy & 15.2\% & 30.5\% & 16.5\% & 37.6\% & 24.6\% & 7.9\% & 19.0\% & 48.5\% \\ 
        Naturalness of movement & 12.8\% & 14.0\% & 18.6\% & 54.5\% & 19.0\%  & 6.3\% & 7.1\% & 67.6\% \\   
        Image quality & 8.3\% & 7.2\% & 14.3\% & 70.0\% & 22.8\% & 11.1\% & 16.7\% & 49.7\% \\
        \bottomrule
    \end{tabular}
\end{table*}

\begin{figure*}
  \centering
  \begin{subfigure}{0.3\linewidth}
    \includegraphics[width=\linewidth]{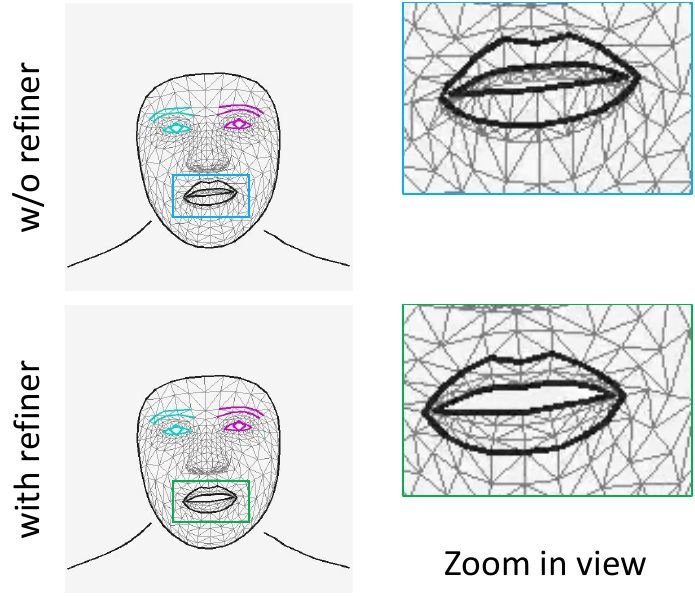}
    \caption{Effectiveness of FaCo-Net.}
    \label{fig:ablation_others-a}
  \end{subfigure}
  \hfill
  \begin{subfigure}{0.66\linewidth}
    \includegraphics[width=\linewidth]{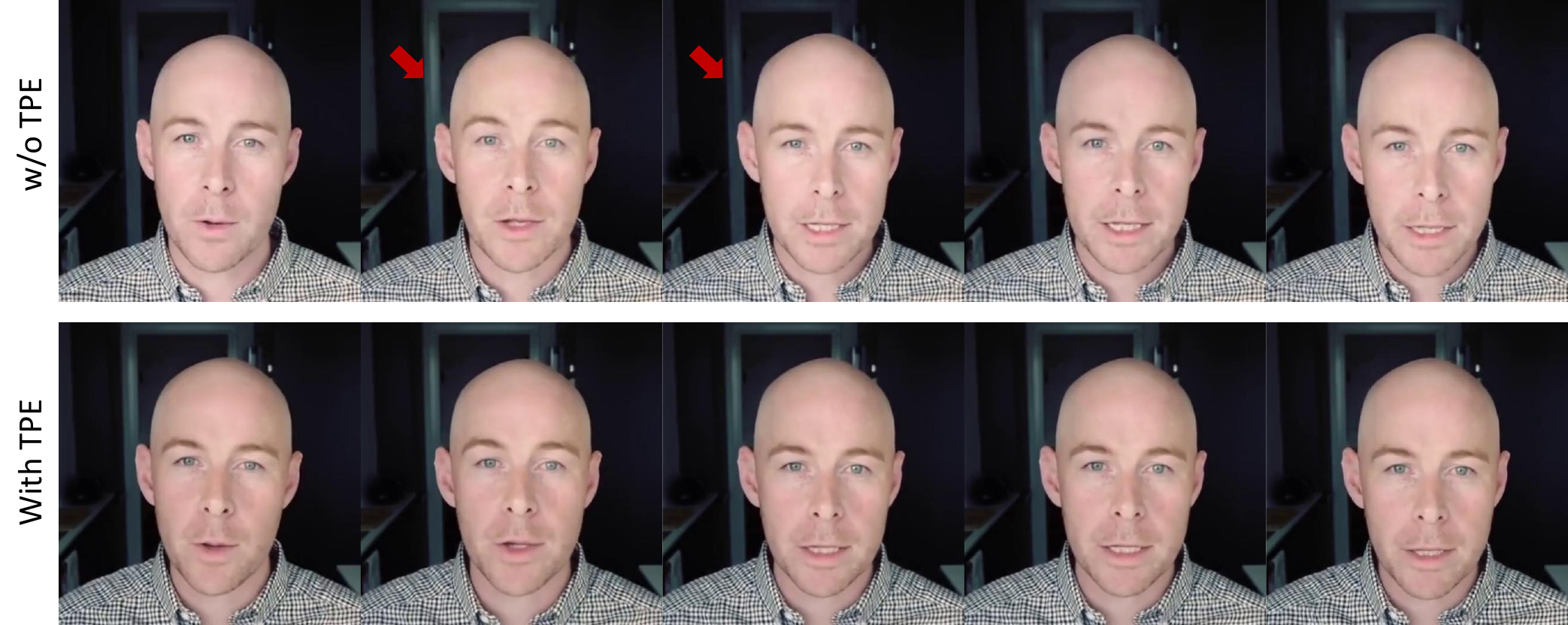}
    \caption{Effectiveness of TPE.}
    \label{fig:ablation_others-b}
  \end{subfigure}
  \caption{Ablation studies on FaCo-Net (a) and temporal positional encoding (b).}
  \label{fig:ablation_others}
\end{figure*}

\subsection{Qualitative Evaluation}
\noindent \textbf{User Study.} We conduct user studies with 20 attendees on 30 videos generated by ours and the other methods. The driving audio is selected from four different languages: English, Chinese, Japanese, and German. The videos are generated across 5 subjects. Each participant is asked to select the best generated talking-portrait videos based on three major aspects: lip synchronization accuracy, the naturalness of movements including head movement, eye blinking, and upper body movement, and the video quality of the generated portrait. 
We collect the voting results and calculate the best-voting percentage of each method. The statistics are reported in \cref{tab:user_study}. Overall, users prefer our results on lip synchronization, the naturalness of portrait, and video quality, indicating the effectiveness of the proposed method. 

\noindent \textbf{Qualitative comparison.} 
\cref{fig:vis_cmp} demonstrates the visual comparison among different methods.
The results from LSP~\cite{lu2021LSP} have some warping effects without 3D consistency. Wav2Lip~\cite{prajwal2020lip} can generate accurate mouth motions. However, their mouth areas usually have blurry boundaries and artifacts, which make the video unnatural. The results from AD-NeRF~\cite{guo2021adnerf} have blurry boundaries of shoulders. SadTalker~\cite{zhang2022sadtalker} may suffer from out of sync. GeneFace~\cite{ye2023geneface} has obvious artifacts on the neck region. Compared to these methods, our system generates portrait videos with overall high-quality and natural mouth movements. 

\noindent  \textbf{Diverse outputs.}
\cref{fig:multimodal} shows the diverse rendered videos that are driven by the same audio. These videos have different head poses, eye-blinking, and upper bodies while sharing the same mouth structures. These results demonstrate that our MODA network is able to generate vivid and diverse talking portrait videos.

\subsection{Ablation Study and Performance Analysis}

We conduct ablation studies on dual-attention in MODA, FaCo-Net, and TPE in portrait renderer.

\noindent \textbf{Dual-attention module.} We choose to 1) replace DualAttn with a multi-layer LSTM block~\cite{LSTM}; 2) remove the specific attention branch and 3) remove the multimodal attention branch to evaluate the effectiveness of the dual-attention module. Numerical results on LSP test set are reported in \cref{tab:us_dual_attn}. Using LSTM block cannot generate multimodal results and the diverse score (here we use the variance of the generated facial landmarks) drops to 0. When removing the specific attention branch from the dual attention block, the MODA generates the over-smoothed lip movement, which may be out of lip synchronization and has large LMD and LMD-$v$ errors.

\begin{table}[!t]
    \centering
    \caption{Ablation study on MODA. Removing dual attention or replacing it with LSTM block has negative effects.}
    \label{tab:us_dual_attn}
    \vspace{-3mm}
    \setlength{\tabcolsep}{8.3pt}
    \small
    \begin{tabular}{lccc}
        \toprule
        Method & LMD-$v$ & LMD & Diverse \\ 
        \midrule
        replace with LSTM & 2.49 & 2.79 & 0 \\ 
        w/o multimodal attention & 3.01 & 2.81 & 0  \\ 
        w/o specific attention & 1.80 & 1.55 & 1.70  \\ 
        \midrule
        Final & \textbf{1.79} & \textbf{1.50} & 1.57 \\
        \bottomrule
    \end{tabular}
\end{table}

\noindent \textbf{FaCo-Net.}
The FaCo-Net aims to generate natural and consistent representations for our portrait renderer. We carry out an ablation study on it by removing this stage and directly replacing the eye landmarks and mouth landmarks with facial dense landmarks. \cref{fig:ablation_others-a} shows that condition images without FaCo-Net contain incorrect connections in the lip area and lose face details, leading to low SSIM ($0.871 \to 0.843$), PSNR ($24.77 \to 21.96$) and NIQE ($5.55 \to 6.71$) rendered images (as in \cref{tab:us_refiner}). These results consistently prove the effectiveness of FaCo-Net. 

\begin{table}[!t]
    \centering
    \caption{Ablation study on FaCo-Net.}
    \label{tab:us_refiner}
    \small
    \setlength{\tabcolsep}{8.3pt}
    \begin{tabular}{lccc}
        \toprule
        Method & SSIM $\uparrow$ & PSNR $\uparrow$ & NIQE $\downarrow$ \\ 
        \midrule 
        w/o FaCo-Net Net & 0.843 & 21.96 & 6.71  \\ 
        Final & \textbf{0.871} & \textbf{24.77} & \textbf{5.55}  \\ 
        \bottomrule
    \end{tabular}
\end{table}

\noindent \textbf{Temporally positional encoding.}
We adopt the temporal consistency metric to measure to evaluate the frame-wise consistency (TCM~\cite{2020Unsupervised}) of the generated videos. Specifically, the TCM is defined as 
\begin{equation}
    \text{TCM} = \frac{1}{T}\sum^{T}_{t} \exp{(-\frac{|| O_t - \text{warp}(O_{t-1}) ||^2}{|| V_t - \text{warp}(V_{t-1}) ||^2}-1)},
\end{equation}
where $O_t$ and $V_t$ represent the $t^{th}$ frame in the referenced video (O) and generated video (V), respectively. $\text{warp}(\cdot)$ is the warping function using the optical flow~\cite{Berthold1981Determining}. The 2-norm of a matrix $|| \cdot ||$ is the sum of squares of its elements. Through this equation, the generated video (V) is encouraged to be temporally consistent according to variations in the reference video (O).
\cref{fig:ablation_others-b} demonstrates the comparison of video sequences with/without TPE. We find TPE can stabilize video synthesis, especially when training videos with changing backgrounds. Numerical results in \cref{tab:us_tpe} also show that TPE can increase TCM score.

\begin{table}[!t]
    \setlength{\tabcolsep}{24 pt}
    \centering
    \caption{Ablation study on TPE. Higher is better.}
    \label{tab:us_tpe}
    \small
    \begin{tabular}{cc}
        \toprule
        Method & TCM $\uparrow$ \\ 
        \midrule
        Renderer w/o TPE & 0.63  \\ 
        Renderer with TPE &  \textbf{0.71}  \\ 
        \bottomrule
    \end{tabular}
    \vspace{-0.5cm}
\end{table}

\section{Discussions and Conclusions}
\label{sec:conc}
We present a deep learning approach for synthesizing multimodal photorealistic talking-portrait animation from audio streams. Our method can render multiple personalized talking styles with arbitrary audio. Our system contains three stages, \ie, MODA, FaCo-Net, and a high-fidelity portrait renderer with temporal guidance. The first stage generates lip motion, head motion, eye blinking, and torso motion with a unified network. This network adopts a dual-attention mechanism and is able to generate diverse talking-portrait representations with correct lip synchronization. The second stage generates fine-grained facial dense landmarks powered by generated lip motion and eye blinking. Finally, we generate the intermediate representations for our temporal-guided renderer to synthesize both high-fidelity and stable talk-portrait videos. Experimental results and user studies show the superiority of our method. Analytical experiments have also verified different parts of our system.

\textit{Limitations and future work.}
While our approach achieves impressive results in a wide variety of scenarios, there still exist several limitations. Similar to most deep learning-based methods, our method cannot generalize well on unseen subjects or extremely out-of-domain audio. It may require fine-tuning the renderer for new avatars. We also looking forward to future work to find a person-invariant renderer to achieve high-quality synthesis without additional finetuning. 

{\small
\bibliographystyle{ieee_fullname}
\bibliography{egbib}
}

\newpage
\appendix

\section{Implementation details of FaCo-Net}

\begin{figure*}[htbp]
    \centering
    \includegraphics[width=\linewidth]{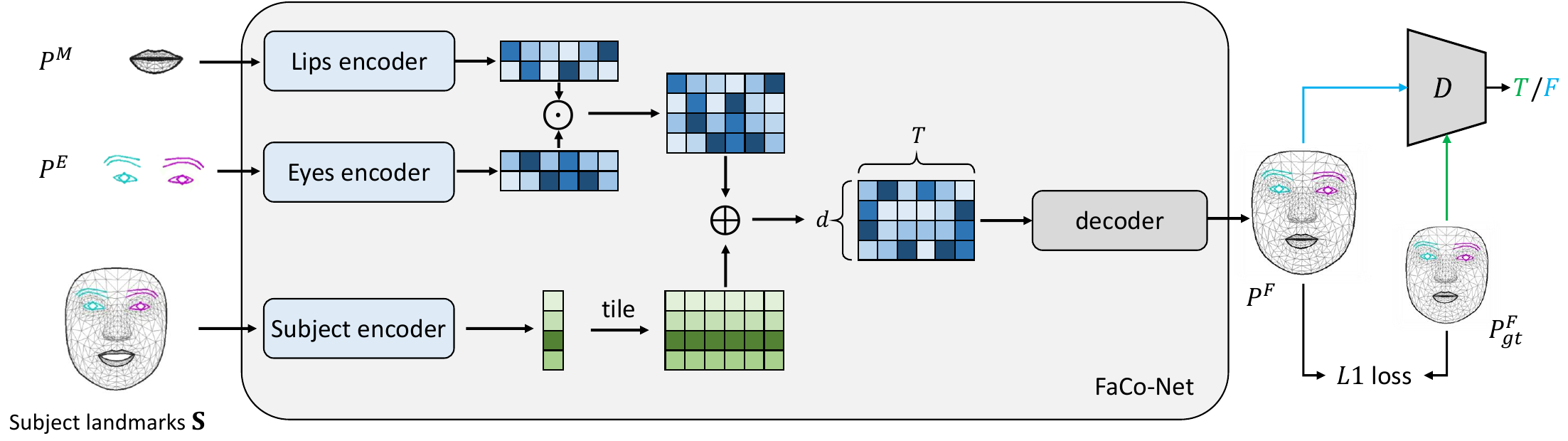}
    \caption{Architecture of FaCo-Net.}
    \label{fig:supp_FaCoNet}
\end{figure*}

This section introduces the network structure and implementation details of FaCo-Net. 
As shown in \cref{fig:supp_FaCoNet}, given the mouth keypoints $P^M$, eye keypoints $P^E$, and the landmark of the subject $\mathbf{S}$, FaCo-Net aims to generate facial details $P^F$ that are consistent with the input mouth and eye keypoints and keeps the target subject style. The computational process is $P^F = \text{FaCo-Net}(\mathbf{S}, P^M, P^E)$. Firstly, FaCo-Net uses three encoders to encode the mouth features, eye features, and target style features, respectively. %
Each encoder is implemented using an MLP. Mathematically, the calculation process is as follows:

\begin{equation}
    P^F = \Psi_c((\mathbf{p}_m \odot \mathbf{p}_e)\oplus \mathbf{p}_f),  
\end{equation}
where $\mathbf{p}_m = \Psi^M(P^M)$, $\mathbf{p}_e = \Psi^E(P^E)$, $\mathbf{p}_f = \text{tile}(\Psi^F(\mathbf{S}))$. $\Psi^M, \Psi^E, \Psi^S$ are encoders for mouth keypoints, eye keypoints, and landmarks of the subject, respectively. $\Psi_c$ is the decoder of FaCo-Net. Afterward, the intermediate feature is decoded by an MLP-based decoder to obtain the overall facial keypoints. In order to make the generated facial keypoints have rich details and avoid over-smoothing, we add GAN loss as one of the objective functions. Specifically, the overall objective function of FaCo-Net is defined in Eq.(13) of the main paper. We use a discriminator implemented by an MLP to calculate the GAN loss. The loss function of the discriminator is Eq.(12) of the main paper.
During the training stage, FaCo-Net and the discriminator are trained alternately. In the testing process, the discriminator will be discarded, and only FaCo-Net will be used for inference.

\section{Architecture and loss functions of portrait renderer}

\begin{figure*}[!th]
    \centering
    \includegraphics[width=\linewidth]{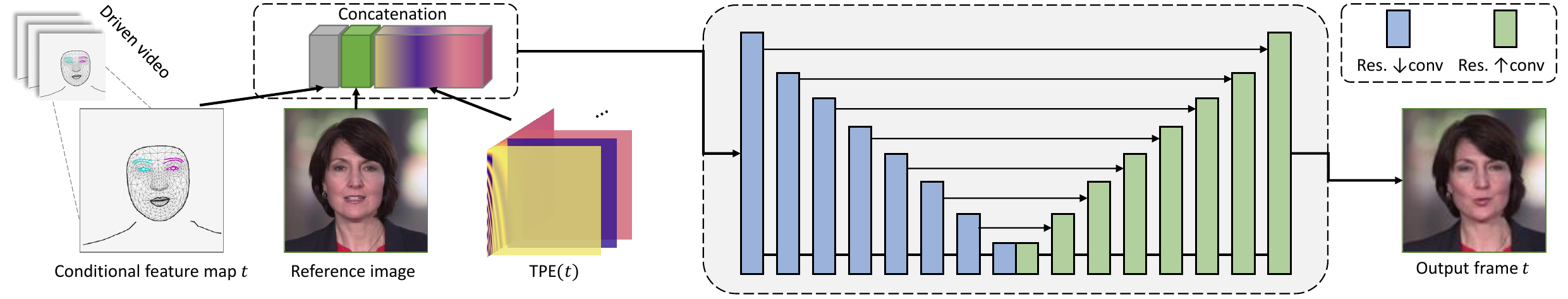}
    \caption{Architecture of portrait renderer.}
    \label{fig:supp_render}
\end{figure*}

The purpose of portrait rendering is to generate high-definition and realistic portrait videos. \cref{fig:supp_render} shows the network architecture of our portrait renderer. Firstly, the network concatenates and fuses the conditional feature map of the $t$-th frame, a reference image, and the TPE at the $t$-th moment in the channel dimension. The generator of the network consists of a U-Net with skip connections. In detail, the network is an 8-layer UNet-like~\cite{2020Unsupervised, lu2021LSP} convolutional neural network with skip connections in each resolution layer. The resolution of each layer is $(256^2, 128^2, 64^2, 32^2, 16^2, 8^2, 4^2, 2^2)$ and the corresponding numbers of feature channels are $(64, 128, 256, 512, 512, 512, 512, 512)$. Each encoder layer consists of one convolution (stride 2) and one residual block. The decoder of the portrait renderer has a structure that mirrors the encoder, which consists of 8 residual convolutional modules with upsampling layers. There are skip connections between each encoder layer and its corresponding decoder layer to better propagate feature information across different levels.

The training process of portrait renderer follows a generative adversarial training strategy. We use a discriminator D with a multi-scale PatchGAN architecture. The purpose of discriminator $D$ is to classify the results generated by generator $G$ as fake and the real images as real. Specifically, we use the LSGAN loss as the adversarial loss to optimize discriminator $D$:

\begin{equation}
    \mathcal{L}_{GAN}(D) = (p^* - 1)^2 + p^2,
\end{equation}
where $p^*, p$ represents the classification result of the discriminator when given a real image $I_t^*$ and an image $I_t$ generated by the generator, respectively. For the generator ($G$)'s loss function, we draw on~\cite{lu2021LSP} and incorporate color loss, mouth loss, perceptual loss, and feature matching loss to further optimize the generator's output. The generator's loss is defined as:

\begin{equation}
    \mathcal{L}_G = \mathcal{L}_{GAN}(G) + \lambda_C\mathcal{L}_C + \lambda_M\mathcal{L}_M + \lambda_P\mathcal{L}_P + \lambda_{FM}\mathcal{L}_{FM},
\end{equation}
where $\mathcal{L}_{GAN}(G)=(p - 1)^2$ is the adversarial loss, $\mathcal{L}_C$ is the color loss, $\mathcal{L}_M$ is the mouth loss, $\mathcal{L}_P$ is the perceptual loss, and $\mathcal{L}_{FM}$ is the feature matching loss. In our experiments, the hyper-parameters are set based on empirical values (50, 100, 10, 1). For the color consistency loss, we use L1 distance, \ie, $\mathcal{L}_C = |I_t - I_t^*|_1$. To enhance the network's ability to generate mouth details, we use a mouth mask to compute the mouth loss, $\mathcal{L}_M = |M I_t - M I_t^*|_1$, where $M$ is the mouth mask. For the perceptual loss, we use VGG19 to extract perceptual features and minimize the $L1$ distance between the generated image features and the ground truth image features. To improve the stability of the training process, we also add the feature matching loss $\mathcal{L}_{FM}=\sum^L_l |y - y^*|$ in the overall objective function, where $L$ is the number of spatial layers in the discriminator. $y$ and $y^*$ are the intermediate predictions in $D$ for the generated image and ground truth, respectively.

\begin{figure*}
    \centering
    \includegraphics[width=\linewidth]{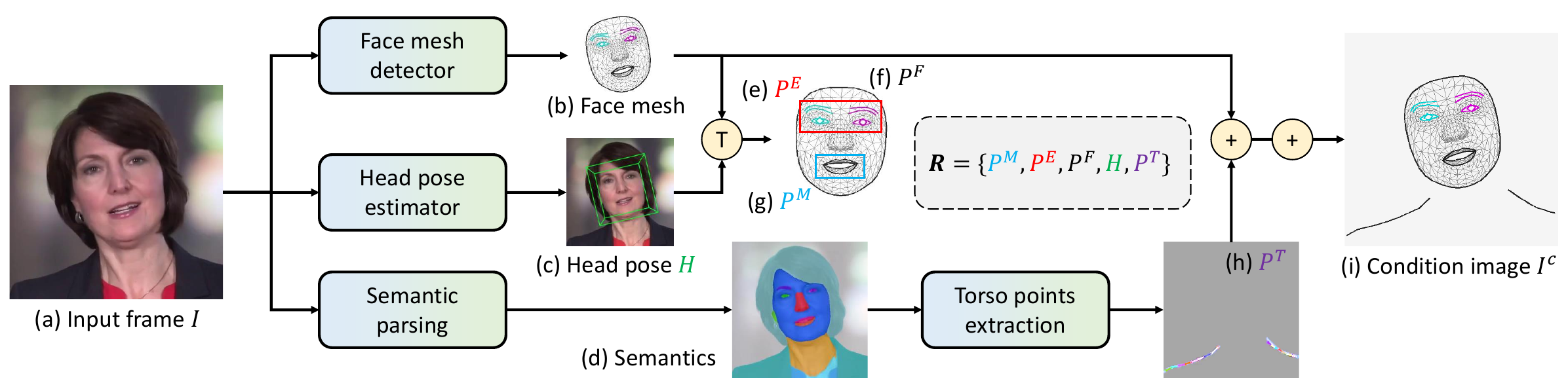}
    \caption{Pipeline of data pre-processing.}
    \label{fig:supp_data_proc_pipeline}
\end{figure*}

\begin{figure*}
    \centering
    \includegraphics[width=\linewidth]{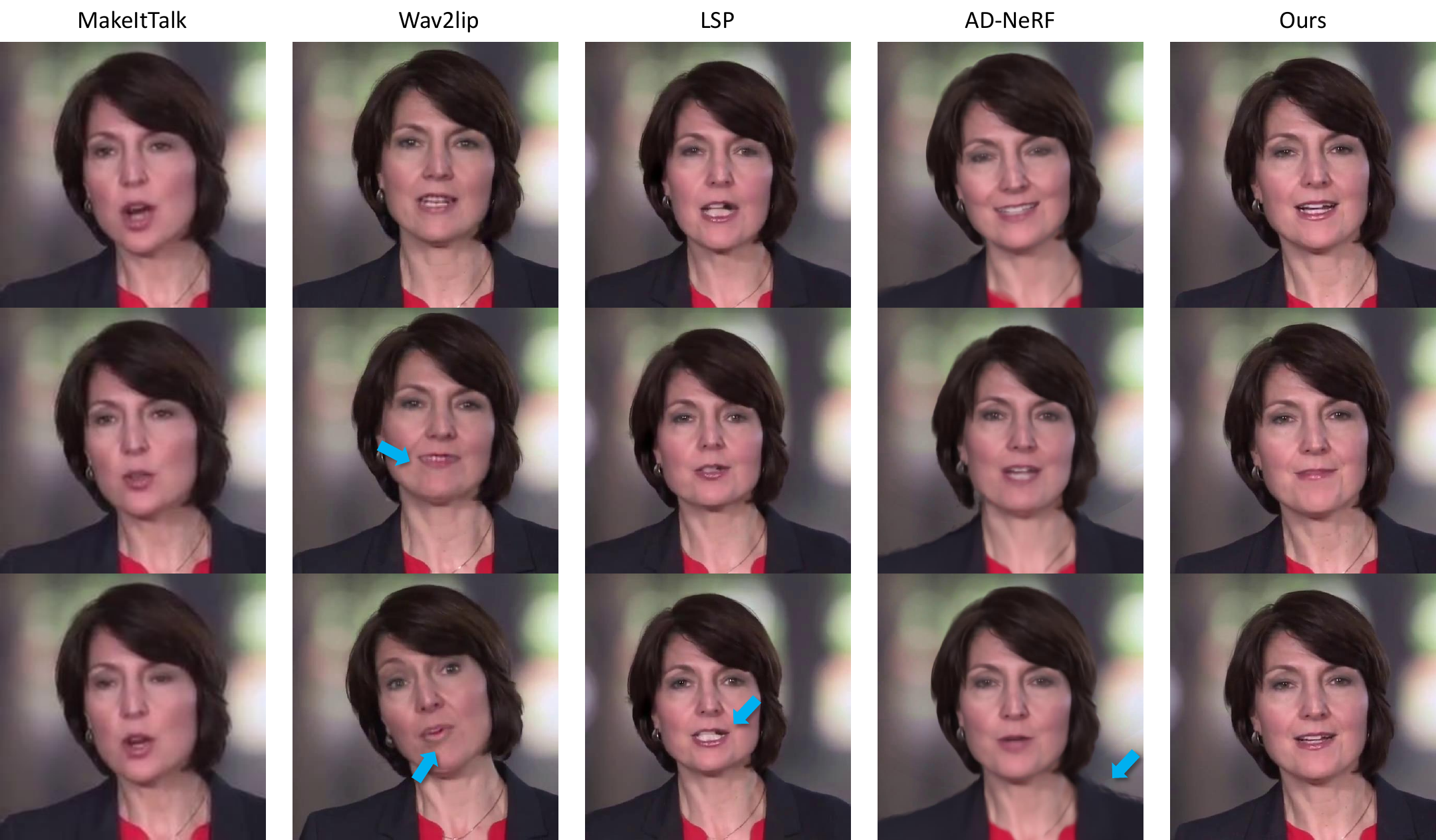}
    \caption{Additional comparisons among different methods.}
    \label{fig:supp_visual_Cmp}
\end{figure*}

\section{Data pro-processing pipeline}

The purpose of data pre-processing is to extract facial keypoints, head pose, and other information from videos to train networks at different stages. The data pre-processing process is shown in \cref{fig:supp_data_proc_pipeline}. For the input video frame $I$, we first use Mediapipe\footnote{\url{https://google.github.io/mediapipe/}} to extract 478 3D facial keypoints (b). Then we use WHENet~\cite{2020WHENet} to estimate the head pose H of the person. By utilizing the head pose, we align the facial keypoints with a rigid transformation (\ie, \textcircled{\scriptsize{T}} in \cref{fig:supp_data_proc_pipeline}) to standard space to align the facial keypoints of different frames in the video, which are denoted as $P^F$. We extract the keypoints in the eye area and mouth area of $P^F$ as the ground truth for training MODA. The eye keypoints $P^E$ and mouth keypoints $P^M$ are illustrated at (e) and (d) in \cref{fig:supp_data_proc_pipeline}, respectively.

To accurately extract shoulder information as a condition for the torso, we design a semantic-guided 3D torso points estimation method. Specifically, we first use BiSeNet~\cite{yu2021bisenet} to segment semantic information (d) from the image $I$. Furthermore, we design a torso points extraction algorithm to estimate key points information for the upper body. The algorithm consists of the following steps:
\begin{enumerate}
    \item We first calculate the semantic boundary of the upper body by computing the boundary between the upper body semantics and the background/hair semantics.
    \item Then, we use morphological operations on the semantic boundary to expand its range, and we extract key points from the semantic contour using a polygon fitting algorithm.
    \item Next, we use a $k$-nearest neighbors algorithm to constrain the number of key points for each side of the shoulder. $k$ is set to 9 in our experiments.
    \item After obtaining the 2D key points of the torso, we use the average depth information of the face mesh (b) as the depth information of the torso keypoints. 
\end{enumerate}
The visualization result of the extracted body keypoints is shown in \cref{fig:supp_data_proc_pipeline}(h). By adding the face mesh (b) and upper body key points $P^T$ (h) projected onto the image coordinate, we obtain the condition image $I^c$ (i) of the portrait image.

For the training of the proposed system, given a reference image of a subject $I_r$, we extract the face mesh obtained from a face mesh detector as the style $\mathbf{S}$. The audio information $\mathbf{A}$ and $\mathbf{S}$ are used as input, $P^M, P^E, H, P^T$ in \cref{fig:supp_data_proc_pipeline} are used as the target, to train the first stage of MODA. The goal of the second stage, FaCo-Net, is to learn the mapping from $S, P^E, P^M$ to $P^F$, so that the generated $P^F$ contains rich details. Finally, the condition image $I^c$, input image $I$, and reference image $I_r$ are combined to form the training data for the portrait renderer.

\section{Additional experimental results}

\subsection{Additional visual comparison results}
In this section, we provide additional visual comparison results among different methods in
\cref{fig:supp_visual_Cmp}. MakeItTalk~\cite{zhou2020MakeItTalk} generates low-resolution videos without head/torso motions. The results from LSP~\cite{lu2021LSP} have some warping effects and are not 3D consistent. Wav2Lip~\cite{prajwal2020lip} can generate accurate mouth motions. However, their mouth areas usually have blurry boundaries and artifacts, which make the video unnatural. The results of AD-NeRF~\cite{guo2021adnerf} have blurry boundaries around shoulders, and the relative movements between the head and torso are unnatural. Compared to other baselines, our system generates portrait videos with correct lip-sync, natural movements, and high visual quality.

\begin{table*}[!t]
    \centering
    \caption{Running time comparisons between the proposed method and other methods.}
    \label{tab:supp_running_time}
    \small
    \setlength{\tabcolsep}{8.3pt}
    \begin{tabular}{lcccccc}
        \toprule
        ~ & \multicolumn{3}{c}{Training time} & \multicolumn{3}{c}{Inference time}\\
        
        \cmidrule(lr){2-4} \cmidrule(lr){5-7}
        Method & 1 subject & 2 subjects & 3 subjects & 5s audio & 10s audio & 30s audio  \\ 
        \midrule
        LSP~\cite{lu2021LSP} & $\sim$ \textbf{14h} & $\sim$ 30h & $\sim$ 50h & 15s & 26s & 70s \\ 
        AD-NeRF~\cite{guo2021adnerf} & $\sim$ 70h & $\sim$ 145h & $\sim$ 220h & $\sim$ 11min & $\sim$ 25min & $\sim$ 80min \\ 
        GeneFace~\cite{ye2023geneface} & $\sim$ 85h & $\sim$ 150h & $\sim$ 230h & $\sim$ 26min & $\sim$ 92min & $\sim$ 270min \\ 
        \midrule
        MODA (Ours) & $\sim$ 15h & $\sim$ \textbf{17h} & $\sim$ \textbf{20h} & \textbf{12s} & \textbf{25s}  & \textbf{62s} \\ 
        \bottomrule
    \end{tabular}
\end{table*}

\subsection{Running time comparisons}
In this section, we provide running time comparisons among different methods that can generate high-fidelity videos. All models are trained and tested under the same condition (\ie, a single RTX 3090 GPU). Results are reported at \cref{tab:supp_running_time}. 
Since the compared methods require training the network separately for each subject, their training time increases proportionally with the number of subjects. Our method, on the other hand, can generalize across multiple individuals and therefore can be trained simultaneously on multiple subjects, resulting in significant time reduction, especially as the number of training subjects increases (\eg, $\mathbf{2.5}\times, \mathbf{11.5}\times$ faster than LSP and GeneFace under 3 subjects).
During the inference stage, LSP needs to use different networks to generate mouth movements and head movements separately, while our method can generate multiple features to drive the portrait through mapping once, resulting in faster overall inference time. Both AD-NeRF and GeneFace require the use of NeRF to render each frame, which significantly slows down the inference speed.
Overall, our method achieves faster training and inference speed, demonstrating the superiority of our proposed approach.

\end{document}